\documentclass[twoside,11pt]{article}

\usepackage{automl2019}
\usepackage{booktabs}
\usepackage{float}

\makeatletter
\newcommand{\printfnsymbol}[1]{%
  \textsuperscript{\@fnsymbol{#1}}%
}
\makeatother

\jmlrheading{P. Gijsbers, E. LeDell, J. Thomas, S. Poirier, B. Bischl and J. Vanschoren}

\ShortHeadings{AutoML Benchmark}{Gijsbers, LeDell, Thomas, Poirier, Bischl and Vanschoren}
\firstpageno{1}

\begin{document}

\title{An Open Source AutoML Benchmark}


\author{\name Pieter Gijsbers\textsuperscript{1}\textsuperscript{*} \email p.gijsbers@tue.nl 
       \AND
       \name Erin LeDell\textsuperscript{2}\textsuperscript{*} \email erin@h2o.ai 
       \AND
       \name Janek Thomas\textsuperscript{3}\textsuperscript{*} \email janek.thomas@stat.uni-muenchen.de 
       \AND
       \name S\'ebastien Poirier\textsuperscript{2}\textsuperscript{*} \email sebastien@h2o.ai 
       \AND       
       \name Bernd Bischl\textsuperscript{3} \email bernd.bischl@stat.uni-muenchen.de
       \AND
       \name Joaquin Vanschoren\textsuperscript{1} \email j.vanschoren@tue.nl \\
       \textsuperscript{*} \textit{The four authors contributed equally to the paper.} \\
       \addr \textsuperscript{1} Eindhoven University of Technology, Netherlands \\
       \addr \textsuperscript{2} H2O.ai, United States \\
       \addr \textsuperscript{3} Ludwig-Maximilians-University Munich, Germany 
       }

\maketitle

\begin{abstract}
In recent years, an active field of research has developed around automated machine learning (AutoML).
Unfortunately, comparing different AutoML systems is hard and often done incorrectly.
We introduce an open, ongoing, and extensible benchmark framework which follows best practices and avoids common mistakes.
The framework is open-source, uses public datasets and has a website with up-to-date results.
We use the framework to conduct a thorough comparison of 4 AutoML systems across 39 datasets and analyze the results.

\end{abstract}

\section{Introduction}
Designing and tuning machine learning systems is a labor and time intensive task which requires extensive expertise. The field of automated machine learning (AutoML) is focused on automating this task. AutoML tools allow novice users to create useful machine learning models, while experts can use them to free up valuable time for other tasks. In recent years, many approaches have been developed for building and optimizing model learning pipelines, or building and optimizing deep neural networks. This paper focuses on the former.


\subsubsection*{The Need for Standardized Benchmarks}
There is no universally best AutoML approach. Hence, we need comparisons to help practitioners select the right tools, and provide objective feedback to the research community.
Unfortunately, many current comparisons are lacking in several ways. The selection of datasets is often too limited in scope, typically reusing the same (mostly small) datasets used in comparisons many years ago~\citep{AutoWEKA1}. This increases the possibility of overfitting on a specific set of datasets, creates a bias towards older datasets rather than current challenges, and may fail to show individual tool's strengths or weaknesses. Authors may even knowingly or unknowingly select datasets on which current systems perform well. Finally, `rival' methods may not have been run correctly, for instance by misunderstanding memory management and/or using insufficient compute resources ~\citep{DBLP:journals/corr/abs-1808-06492}.


\subsubsection*{A New Hope}
In this work, we present an open, extensible and ongoing AutoML benchmark to address these problems.
The benchmark is completely \textit{open source}\footnote{\url{https://github.com/openml/automlbenchmark/}}, and allows anyone to \textit{extend} it by adding or updating AutoML systems through pull requests. 
Finally, it is \emph{ongoing} because we will update it with new benchmark datasets, run the experiments again when AutoML tools have substantial version updates.
The benchmark is accompanied by a website which will show the latest results and other information.\footnote{\url{https://openml.github.io/automlbenchmark/}} 


\section{Related Literature \label{sec:literature}}
AutoML methods differ in their optimization method (e.g. Bayesian Optimization or Genetic Programming), the pipelines they generate (e.g. with or without fixed structure), the library of algorithms they select from, whether they use meta-learning to learn from runs on prior datasets, or whether they perform post-processing (e.g. ensemble construction).


\begin{table}[t]
    \begin{tabular}{ l r r r r }
    \toprule
     Tool & Back-end & Optimization & Meta-learning & Post-processing\\ 
    \midrule
     Auto-WEKA & WEKA & Bayesian  & - & - \\
     auto-sklearn & scikit-learn& Bayesian  & warm-start & ensemble selection\\  
     TPOT & scikit-learn & Genetic Programming  & - & - \\
     H2O AutoML & H2O & Random Search  & - & stacked ensembles \\
     
     \bottomrule
    \end{tabular}
    \caption{Simplified comparison of a selection of AutoML tools.}
    \label{tab:automltools}
\end{table}

Table~\ref{tab:automltools} shows a simplified comparison of the AutoML tools compared in this paper. The first prominent AutoML tool was \textbf{Auto-WEKA}~\citep{AutoWEKA1}, which used Bayesian optimization to select and tune the algorithms in a machine learning pipeline based on WEKA~\citep{WEKA}. 
\textbf{auto-sklearn}~\citep{feurer2015efficient} did the same using scikit-learn~\citep{pedregosa2011scikit} and added meta-learning to \textit{warm-start} the search with the best pipelines on similar datasets, as well as ensemble construction. \textbf{TPOT}~\citep{TPOT} optimizes scikit-learn pipelines via genetic programming, starting with simple ones and evolving them over generations. Finally, \textbf{H2O AutoML}~\citep{H2OAutoML} optimizes H2O components by stacking the best solutions found by a random search.

Notable omissions from this list include autoxgboost~\citep{autoxgboost}, which leverages Bayesian optimization to optimize gradient boosting models, OBOE~\citep{oboe}, which uses low-rank approximation to predict the best pipelines, ML-Plan~\citep{MLPlan}, which optimizes WEKA-based pipelines using hierarchical planning, and Hyperband~\citep{Hyperband}, a bandit-based approach which aggressively selects configurations based on performance on subsamples of data. We do plan to include these in the next version of the benchmark. In some cases, we ran into technical issues and we are in contact with the authors to resolve them. There are also several AutoML systems which were sadly not yet open sourced at the time of writing, yet we hope that their authors will add them to the benchmark in the near future.


%

Prior efforts to create systematic AutoML benchmarks were sadly obfuscated by memory management and evaluation setup issues and lack strong baselines to interpret the results~\citep{DBLP:journals/corr/abs-1808-06492}.
We instead opted to build an open-source framework in dialogue with the AutoML framework developers to ensure the tools were properly used.
While we evaluate on fewer datasets, they are about 15 times larger on average.
We followed best practices on how to construct and run good machine learning benchmarks \citep{bischl2017openml} and general-purpose algorithm configuration libraries \citep{aslib}, but also extend them since evaluating AutoML systems comes with specific challenges.

\section{Benchmark Design \label{sec:benchmark}}
Each benchmark task consists of a dataset, a metric to optimize, and specific resources to use. We will briefly explain our choice for each. More details are available on the website.
\paragraph{Datasets}
We selected $39$ classification datasets from previous AutoML papers \citep{AutoWEKA1}, competitions \citep{guyon2016brief}, and machine learning benchmarks \citep{bischl2017openml}\footnote{OpenML CC-18: \url{https://openml.github.io/OpenML/benchmark}}, according to a predefined list of criteria.\footnote{For more details, see: \url{https://openml.github.io/automlbenchmark/benchmark_datasets.html}}
To study the differences between AutoML systems, the datasets vary in the number of samples and features by orders of magnitude, and vary in the occurrence of numeric features, categorical features and missing values. We excluded datasets which were too easily solved with AutoML or did not represent typical AutoML scenario's (e.g. artificial datasets). The \textit{current} list of datasets is available on OpenML \citep{vanschoren-sigkdd13a}.\footnote{Study for this benchmark: \url{https://www.openml.org/s/218}} In line with our goal to regularly update the benchmark and avoid overfitting on one static set, we set aside some datasets for inclusion in the near future. We would also like to include even bigger datasets in subsequent versions, as well as regression datasets, currently omitted because of computational constraints.





\paragraph{Performance metrics}
The benchmark can be run with a wide range of measures at the user's discretion. For the results in this paper, area under the receiver operator curve (AUROC) is used for binary classification problems and log loss is used for multi-class classification problems\footnote{We use the implementations provided by scikit-learn 0.20}, mainly because these are insightful, commonly used and supported by most AutoML tools. It is imperative that AutoML system optimize for the same metric they are evaluated on. The measures are estimated with ten-fold cross-validation.

\paragraph{Hardware choice and resource specifications}
To improve reproducibility and extensibilily, we opted to use standard \emph{m5.2xlarge} instances available on Amazon Web Services (AWS).\footnote{$32$~GB memory, $8$~vCPUs (Intel Xeon Platinum 8000 series Skylake-SP processor with a sustained all core Turbo CPU clock speed of up to 3.1 GHz). \url{https://aws.amazon.com/ec2/instance-types/m5/}} These represent commodity level hardware which also keep the cost of running the benchmark down. It is not required to run the benchmark on AWS however, as the benchmark can also be run locally directly on the machine or from a docker container.

\paragraph{Frameworks and their configuration}
We required all AutoML tools to be open source. We selected the current set of tools based on popularity, ease of use and variety of underlying techniques. We do plan to include more tools in future work and encourage all developers to add their AutoML tool to the benchmark framework.
Baseline methods include a constant predictor, which always predicts the class prior, an \textit{untuned} Random Forest\footnote{Unless specified otherwise, Random Forests were built with 2000 estimators and scikit-learn 0.20 defaults.}, and a \textit{tuned} Random Forest for which up to eleven unique values of \emph{max\_features} are evaluated with cross-validation (as time permits), and evaluated by refitting the final model with the optimal \emph{max\_features} values.

The AutoML tools were all used with their default hyperparameter values and search spaces, since most users will use them in this way. The exception are hyperparameters which specified available resources, which were fixed to a specific number of cores, memory and total runtime. This was done to allow a more practical comparison, and because it is practically impossible to homogenize the search spaces for each tool. It is important to realize that no two tools share the exact same search space or optimization method, so from this benchmark no conclusions can be drawn about those.

\paragraph{Open-source, extensible framework structure.}
In developing the benchmark framework, we made sure that it is easily extensible with new frameworks or datasets. New AutoML tools can be easily wrapped and included: each of the current tools required less than 100 lines of wrapper code. Adding a dataset that is hosted on OpenML only takes 3 lines of code.
New additions are not evaluated automatically.

\paragraph{Meta-learning}
AutoML frameworks may use meta-learning to learn about good configurations across datasets.
An AutoML framework which used datasets of the benchmark in its meta-learning process will have an unfair advantage on them.
We did not decide how to resolve this issue, and leave it as future work.\footnote{Take a look or join the discussion at \url{https://github.com/openml/automlbenchmark/issues/18}}
From our selection of frameworks, only auto-sklearn uses meta-learning, and we indicate affected datasets in the results.

\section{Results \label{sec:results}}
\begin{figure}[H]
  \centering
  \includegraphics[width=0.5\textwidth]{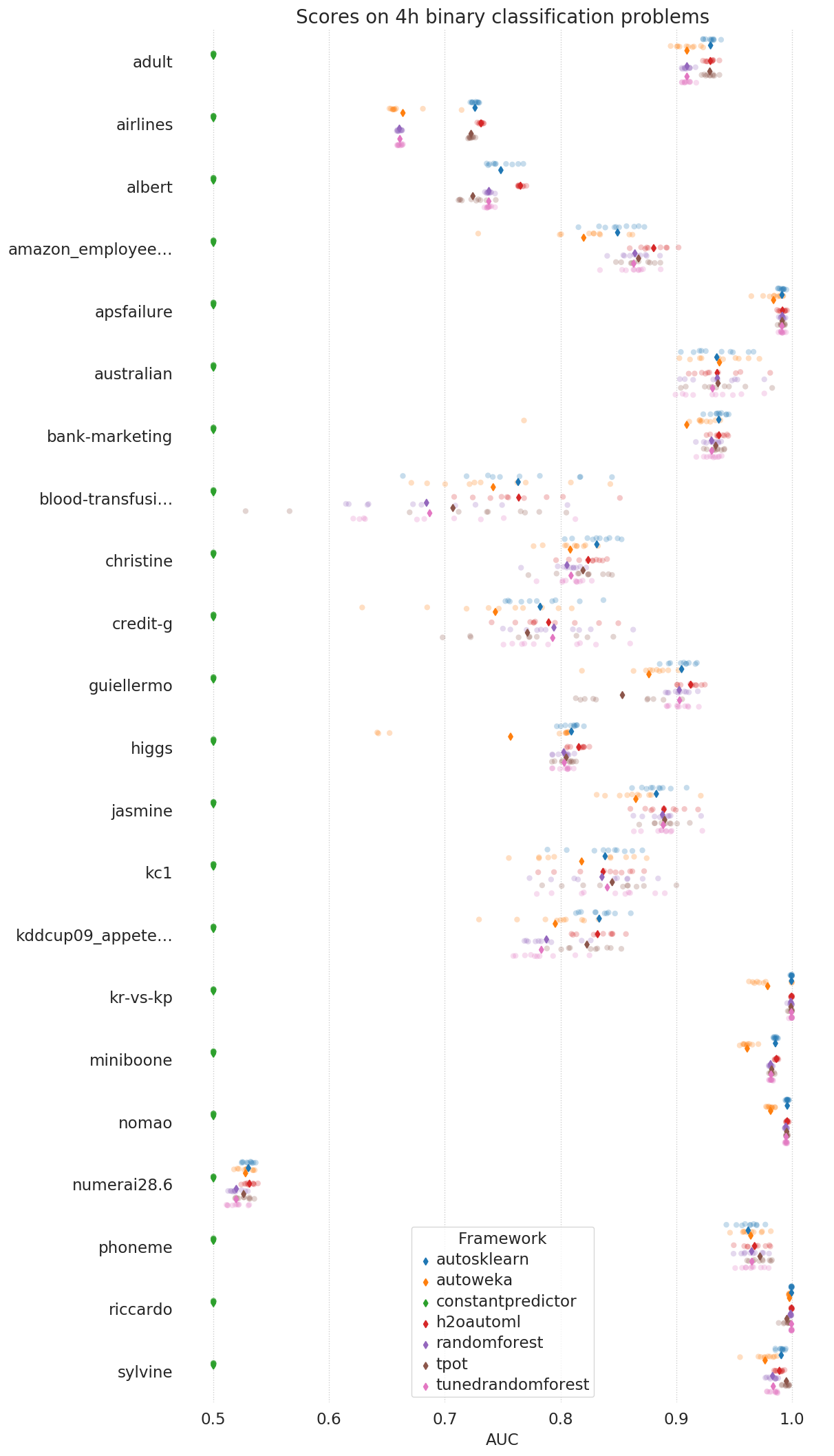}
  \hspace{-0.3cm}
  \includegraphics[width=0.5\textwidth]{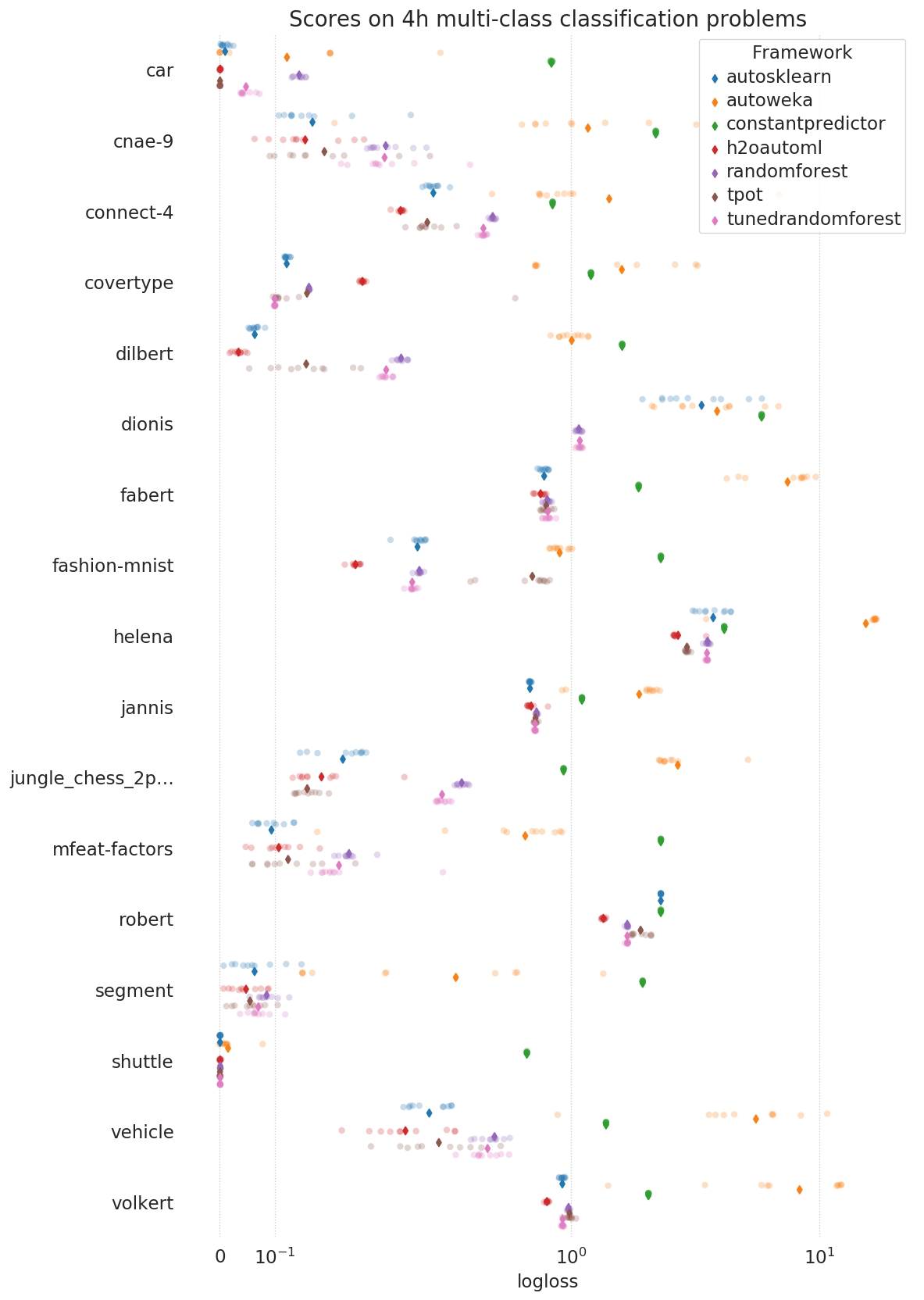}
  \caption{Scores obtained on each dataset by each framework on each of ten folds. On the left are binary classification problems with their AUROC scores, on the right are multi-class classification problems with logloss. Opaque diamonds represent the average score across all folds.}
  \label{fig:results}
\end{figure}

We ran two benchmarks, using a time budget of 1 and 4 hours per fold respectively, for a total of around 8000 hours of computation time.
The 4h `raw' results are visualized in Figure~\ref{fig:results}. Those results are very similar to the 1h ones, bringing only slight score improvements for some frameworks, especially TPOT. Auto-WEKA is showing signs of overfitting when running longer, especially on multi-class problems. There is no AutoML system which consistently outperforms all others. On some datasets, the performance differences can be significant, but on others the AutoML methods are only marginally better than a Random Forest. The variance of the per-fold scores can be quite large. On the `dionis` and `helena` datasets, all frameworks perform worse than a Random Forest. Both have more more than 100, quite unbalanced classes, which seems to be a weak spot for current AutoML techniques, at least under log loss.

Because the scores vary across tasks, we also normalized them such that the constant predictor is $0$ and the tuned random forest is $1$. Shown in Table~\ref{tab:results}, these scores reflect the relative improvement over our strongest baseline, with a score greater than one being better than the strongest baseline. Generally, these scores are very similar across methods, all being relatively close to the tuned random forest baseline. None of the AutoML systems outperforms an untuned random forest across all problems, though in most cases they are better than a tuned random forest. Auto-WEKA has the poorest performance out of the tested AutoML packages under the tested conditions. Note that these results were obtained using a rather generous time budget. We hope to add anytime performance evaluation curves in the future, but this is not yet supported by many of the AutoML tools.

Finally, we can observe that on some datasets, some AutoML tools perform significantly better or worse than others. At the moment, we can't draw clear conclusions about which data properties explain this behavior beyond what we observed above. We aim to study this further by including more datasets.

\begin{table}[H]
\centering
\scriptsize
\begin{tabular}{lrrrrr}
\toprule
Framework: & auto-sklearn & Auto-WEKA & H2O AutoML & RandomForest & TPOT \\
Binary tasks:&&&&&\\
\cmidrule{1-1}
adult & 1.045 & 1.000 & 1.049  & 1.000 & 1.048 \\
airlines & 1.403 & 1.016 & 1.435  & 0.997 & 1.343 \\
albert & 1.009 &  & 1.115  & 1.001 & 0.981 \\
amazon\_employee... & 0.972\textsuperscript{*} & 0.886 & 1.048  & 1.003 & 1.012 \\
apsfailure & 1.000 & 0.985 & 1.001  & 1.000 & 1.001 \\
australian & 1.010 & 1.015 & 0.909  & 1.010 & 1.011 \\
bank-marketing & 1.012 & 0.950 & 1.015  & 1.000 & 1.008 \\
blood-transfusion & 1.495 & 1.379 & 1.532 & 0.985 & 1.149 \\
christine & 1.072 & 0.998 & 1.048  & 0.988 & 1.029 \\
credit-g & 0.970\textsuperscript{*} & 0.829 & 0.991 & 1.004 & 0.924 \\
guiellermo & 1.004 & 0.934 & 1.024  & 0.999 & 0.878 \\
higgs & 1.018\textsuperscript{*} & 0.845 & 1.041  & 0.999 & 1.005 \\
jasmine & 0.987 & 0.939 & 1.001 & 0.998 & 1.004 \\
kc1 & 0.999\textsuperscript{*} & 0.934 & 0.992  & 0.987 & 1.013 \\
kddcup09\_appetency & 1.181\textsuperscript{*} & 1.043 & 1.176  & 1.016 & 1.134 \\
kr-vs-kp & 1.000\textsuperscript{*} & 0.959 & 1.000  & 0.999 & 0.999 \\
miniboone & 1.008 & 0.957 & 1.010  & 0.999 & 1.001 \\
nomao & 1.002 & 0.973 & 1.002 & 1.000 & 1.001 \\
numerai28.6 & 1.679 & 1.544 & 1.730  & 1.042 & 1.428 \\
phoneme & 0.993\textsuperscript{*} & 0.998 & 1.005  & 1.000 & 1.015 \\
riccardo & 1.000 & 0.996 & 1.000  & 0.999 & 0.992 \\
sylvine & 1.013 & 0.985 & 1.011  & 0.999 & 1.023\\
\midrule
Multi-class tasks:&&&&&\\
\cmidrule{1-1}
car & 1.030 & 0.906 & 1.060  & 0.878 & 1.060 \\
cnae-9 & 1.069 & 0.541 & 1.076  & 0.999 & 1.057 \\
connect-4 & 1.184 & -1.565 & 1.409  & 0.954 & 1.276 \\
covertype & 0.976 & -0.361 & 0.856  & 0.944 & 0.933 \\
dilbert & 1.182 & 0.459 & 1.205  & 0.979 & 1.111 \\
dionis & 0.580 & 0.590  &  & 1.002 & \\
fabert & 1.026 & -5.235 & 1.049  & 1.004 & 1.005 \\
fashion-mnist & 0.995 & 0.717 & 1.052  & 0.993 & 0.841 \\
helena & 0.660 & -18.420 & 1.905  & 0.970 & 1.676 \\
jannis & 1.083 & -1.989 & 1.065  & 0.973 & 0.987 \\
jungle\_chess... & 1.299 & -3.309 & 1.235  & 0.933 & 1.459 \\
mfeat-factors & 1.059\textsuperscript{*} & 0.789 & 1.053 & 0.992 & 1.018 \\
robert & -0.001 &  & 1.545  & 1.000 & 0.640 \\
segment & 1.004 & 0.808 & 1.012  & 0.992 & 1.008 \\
shuttle & 1.000 & 0.979 & 1.000  & 1.000 & 1.000 \\
vehicle & 1.102 & -4.630 & 1.166  & 0.986 & 1.099 \\
volkert & 1.002 & -5.585 & 1.111  & 0.954 & 0.945 \\
\bottomrule
\end{tabular}
\footnotetext{Footnote}
\caption{Performance of AutoML frameworks, scaled between a constant class prior predictor (=$0$) and a tuned random forest ($=1$). Missing values mean that no results were returned in time. \textsuperscript{*}: the task was also included in meta-learning models.}
\label{tab:results}
\vspace{-0.5cm}
\end{table}

\section{Conclusion}
We presented a novel benchmark for AutoML frameworks which is open-source, extensible both in terms of AutoML frameworks and tasks, and ongoing, publishing all the latest results online. Current results already highlight several avenues for further AutoML research. On some datasets, none of the frameworks outperforms a Random Forest within 4 hours, and high-dimensional or highly multi-class problems are often challenging. In future work, we will include more frameworks and tasks, especially larger datasets and regression tasks.

\section*{Acknowledgements}
Pieter Gijsbers would like to acknowledge funding by the Data Driven Discovery of Models (D3M) program run by DARPA and the Air Force Research Laboratory.

\vskip 0.2in
\bibliography{ref}


\newpage

\appendix

\end{document}